\documentclass{article}

\usepackage[utf8]{inputenc}

\usepackage{times}
\usepackage{graphicx} % more modern
\usepackage{natbib}
\usepackage{amssymb,amsmath}
\usepackage{multirow}
\usepackage{arydshln} % dashed lines in tables
\usepackage{enumitem}

\usepackage{hyperref}
\usepackage[accepted]{icml2017}

\usepackage[normalem]{ulem}
\definecolor{darkgreen}{rgb}{0.0, 0.5, 0.0}

\def\ODdel#1{\bgroup\markoverwith{\textcolor{darkgreen}{\rule[0.5ex]{2pt}{1pt}}}\ULon{#1}}

\def\VRdel#1{\bgroup\markoverwith{\textcolor{magenta}{\rule[0.5ex]{2pt}{1pt}}}\ULon{#1}}

\def\JNdel#1{\bgroup\markoverwith{\textcolor{red}{\rule[0.5ex]{2pt}{1pt}}}\ULon{#1}}

\newcommand\Tstrut{\rule{0pt}{2.3ex}}       % "top" strut
\newcommand\Bstrut{\rule[-0.9ex]{0pt}{0pt}} % "bottom" strut

\icmltitlerunning{Referenceless Quality Estimation for Natural Language Generation}

\begin{document} 

\twocolumn[
\icmltitle{Referenceless Quality Estimation for Natural Language Generation}



% you can specify symbols, otherwise they are numbered in order
%\icmlsetsymbol{hw}{*}

\begin{icmlauthorlist}
\icmlauthor{Ondřej Dušek}{hw}
\icmlauthor{Jekaterina Novikova}{hw}
\icmlauthor{Verena Rieser}{hw}
\end{icmlauthorlist}

\icmlaffiliation{hw}{Interaction Lab, Heriot-Watt University, Edinburgh, Scotland, UK}

\icmlcorrespondingauthor{Ondřej Dušek}{o.dusek@hw.ac.uk}

\icmlkeywords{neural networks, natural language generation, quality estimation}

\vskip 0.3in
]



\printAffiliationsAndNotice{}  % leave blank if no need to mention equal contribution

\begin{abstract} 
%Abstracts should be a single paragraph, between 4--6 sentences long, ideally.
Traditional automatic evaluation measures for natural language generation (NLG) use costly human-authored references to estimate the quality of a system output. In this paper, we propose a referenceless quality estimation (QE) approach based on recurrent neural networks, which predicts a quality score for a NLG system output by comparing it to the source meaning representation only. 
Our method outperforms traditional metrics and a constant baseline in most respects; 
we also show that synthetic data helps to increase correlation results by 21\% compared to the base system.
Our results are comparable to results obtained in similar QE tasks despite the more challenging setting.
%Our method does not generalise well to out-of-domain data but only needs a small amount of in-domain data to improve performance.
\end{abstract}

\section{Introduction}\label{sec:intro}

Automatic evaluation of natural language generation (NLG) is a complex task due to multiple acceptable outcomes. 
Apart from manual human evaluation, most recent works in NLG are evaluated using word-overlap-based metrics such as BLEU \cite{gkatzia_snapshot_2015}, which compute similarity against gold-standard human references. However, high quality human references are costly to obtain, and for most word-overlap metrics, a minimum of 4 references are needed in order to achieve reliable results \cite{finch2004does}.
%While these metrics estimate the output quality} \VRdel{very well} \VR{reliably} at corpus level \cite{papineni_bleu_2002,coughlin_correlating_2003,galley_deltableu:_2015},
Furthermore, these metrics tend to perform poorly at segment level \cite{lavie_meteor:_2007,chen_systematic_2014,novikova:2017}.

We present a novel approach to assessing NLG output quality without human references, focusing on segment-level (utterance-level) quality assessments.\footnote{%
In our data, a ``segment'' refers to an utterance generated by an NLG system in the context of a human-computer dialogue, typically 1 or 2 sentences in length (see Section~\ref{sec:dataset}).
We estimate the utterance quality without taking the dialogue context into account. Assessing the appropriateness of responses in context is beyond the scope of this paper, see e.g.\ \cite{Liu:EMNLP2016,lowe_towards_2017,curry:2017}.}
We train a recurrent neural network (RNN) to estimate the quality of an NLG output based on comparison with the source meaning representation (MR) only.
This allows to efficiently assess NLG quality not only during system development, but also at runtime, e.g.\ for optimisation, reranking, or compensating low-quality output by rule-based fallback strategies.

To evaluate our method, we use crowdsourced human quality assessments of real system outputs from three different NLG systems on three datasets in two domains.
We also show that adding fabricated data with synthesised errors to the training set increases relative performance by 21\% (as measured by Pearson correlation).

In contrast to recent advances in referenceless quality estimation (QE) in other fields such as machine translation (MT) \cite{bojar_findings_2016} or grammatical error correction 
\cite{napoles_theres_2016}, NLG QE is more challenging because (1) diverse realisations of a single MR are often acceptable (as the MR is typically a limited formal language); (2) human perception of NLG quality is highly variable, e.g. \cite{dethlefs2014cluster};
(3) NLG datasets are costly to obtain and thus small in size.
%\VR{current end-to-end NLG systems include surface realisation as well as content selection, e.g.\ \cite{wen_semantically_2015,dusek_training_2015,lampouras_imitation_2016}}
%\OD{it's rather sentence planning than content selection and in this way NLG is similar to MT, so I'd probably stay away from this argument...}
Despite these difficulties, we achieve promising results -- correlations with human judgements achieved by our system stay in a somewhat lower range than those achieved e.g.\ by state-of-the-art MT QE systems \cite{bojar_findings_2016}, but they significantly outperform word-overlap metrics.
To our knowledge, this is the first work in trainable NLG QE without references.

\begin{figure}[tb]
\centering{
\includegraphics[width=\columnwidth]{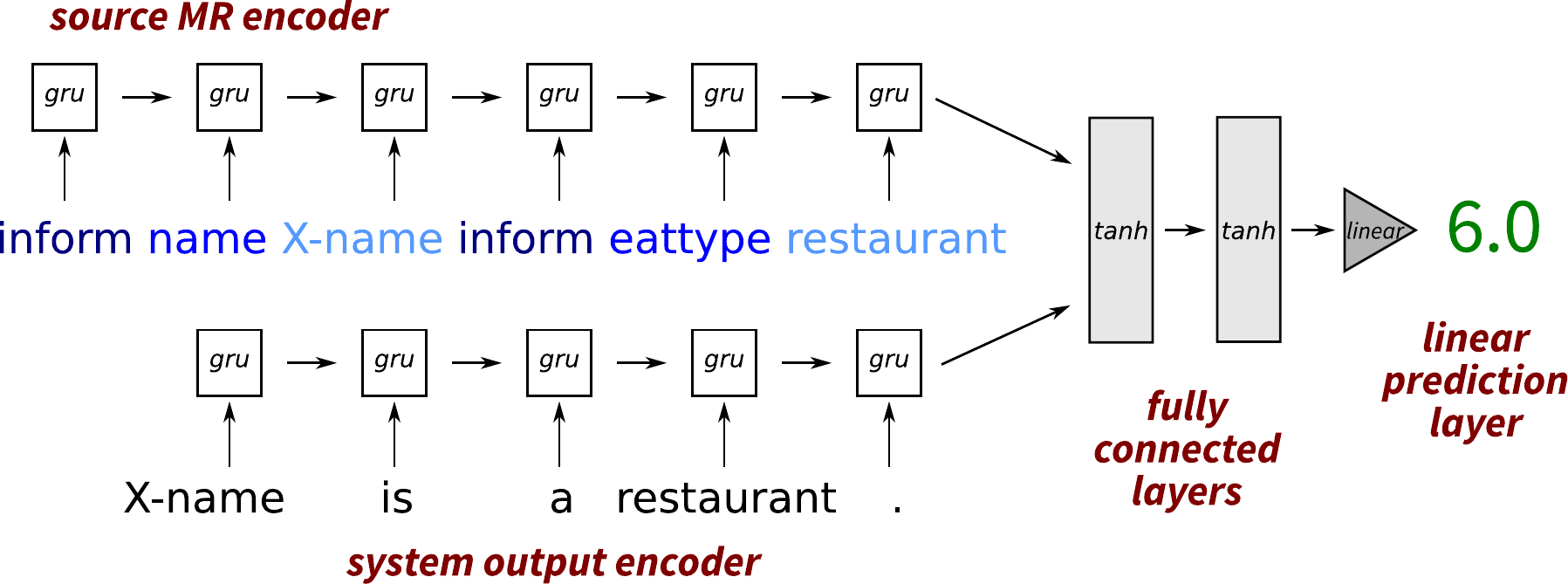}
}
\vspace{-0.3cm}
\caption{The architecture of our referenceless NLG QE model.}\label{fig:model}
\vspace{-0.3cm}
\end{figure}

\vspace{-0.2cm}
\section{Our Model}\label{sec:model}

\vspace{-0.1cm}
We use a simple RNN model based on Gated Recurrent Units (GRUs) \cite{cho_learning_2014}, composed of one GRU encoder each for the source MR and the NLG system output to be rated, followed by fully connected layers operating over the last hidden states of both encoders. The final classification layer is linear and produces the quality assessment as a single floating point number (see Figure~\ref{fig:model}).\footnote{The meaning of this number depends entirely on the training data. In our case, we use a 1--6 Likert scale assessment (see Section~\ref{sec:dataset}), but we could, for instance, use the same network to predict the required number of post-edits, as commonly done in MT (see Section~\ref{sec:related}).}

The model assumes both the source MR and the system output to be sequences of tokens $x^M = \{x_1^M,\dots,x_m^M\}$, $x^S = \{x_1^S,\dots,x_n^S\}$, where each token is represented by its embedding \cite{bengio_neural_2003}.
The GRU encoders encode these sequences left-to-right into sequences of hidden states $\{h^M_t\}_{t=1}^m$, $\{h^S_t\}_{t=1}^n$:
\begin{align}
h^M_x &= \mbox{gru}(x^M_t,h^M_{t-1}) \label{eq:gruM}\\
h^S_x &= \mbox{gru}(x^S_t,h^S_{t-1}) \label{eq:gruS}
\end{align}
The final hidden states $h^M_m$ and $h^S_n$ are then fed to a set of fully connected $\tanh$ layers $z_i, i\in \{0\dots k\}$:
\begin{align}
z_0 &= \tanh (W_0(h^M_m \circ h^S_n)) \label{eq:ff0}\\
z_i &= \tanh (W_k z_{i-1}) \label{eq:ffi}
\end{align}
The final prediction is given by a linear layer:
\begin{equation}
\hat y = W_{k+1} z_k \label{eq:final}
\end{equation}
In (\ref{eq:ff0}--\ref{eq:final}), $W_0\dots W_k$ stand for square weight matrices, $W_{k+1}$ is a weight vector.

The network is trained in a supervised setting by minimising mean square error against human-assigned quality scores on the training data (see Section \ref{sec:experiments}).
Embedding vectors are initialised randomly and learned during training; each token found in the training data is given an embedding dictionary entry.
Dropout \cite{hinton_improving_2012} is applied on the inputs to the GRU encoders for regularisation.

The floating point values predicted by the network are rounded to the precision and clipped to the range found in the training data.\footnote{We use a precision of 0.5 and the 1--6 range (see Section~\ref{sec:dataset}).}

\subsection{Model Variants}

We also experimented with several variants of the model which performed similarly or worse than those presented in Section~\ref{sec:results}.
We list them here for completeness:
\begin{itemize}[itemsep=0pt,topsep=0pt]
\item replacing GRU cells with LSTM \cite{hochreiter_long_1997},
\item using word embeddings pretrained by the word2vec tool \cite{mikolov_efficient_2013} on Google News data,\footnote{We used the model available at \url{https://code.google.com/archive/p/word2vec/}.}
\item using a set of independent binary classifiers, each predicting one of the individual target quality levels (see Section~\ref{sec:dataset}),
\item using an ordered set of binary classifiers trained to predict 1 for NLG outputs above a specified quality level, 0 below it,\footnote{The predicted value was interpolated from classifiers' predictions of the positive class probability.}
\item pretraining the network using a different task (classifying MRs or predicting next word in the sentence). 
\end{itemize}

\section{Experimental Setup}\label{sec:experiments}

In the following, we describe the data we use to evaluate our system, our method for data augmentation, detailed parameters of our model, and evaluation metrics. 

\subsection{Dataset}\label{sec:dataset}

\begin{table}[tb]
\begin{center}
\begin{tabular}{crrrr}\hline
\bf System$\downarrow$ Data$\to$  & BAGEL & SFRest & SFHot & Total\Tstrut\Bstrut \\\hline
LOLS     & 202   & 581    & 398   & 1,181 \\
RNNLG    & -     & 600    & 477   & 1,077 \\
TGen     & 202   & -      & -     &   202 \\\hdashline[0.5pt/2pt]
Total    & 404   & 1,181  & 875   & 2,460\Tstrut\Bstrut \\\hline
\end{tabular}
\end{center}
\caption{Number of ratings from different source datasets and NLG systems in our data.}\label{tab:data-stats}
\end{table}

% used to print raw and quantised fiqures on top of each other, centered
\newcommand{\rawquant}[2]{\multirow{3}{*}{\shortstack[c]{#1 \\ \it #2}}}
\begin{figure*}
\begin{center}\small
\begin{tabular}{llcccccc}\hline
\multicolumn{2}{c}{\bf Instance} & \bf H & \bf B & \bf M & \bf R & \bf C & \bf \ref{li:Ftonly}\Tstrut\Bstrut  \\\hline
\Tstrut\bf MR & inform(name=`la ciccia',area=`bernal heights',price\_range=moderate)\hspace{-0.5cm}  
& \rawquant{5.5}{\phantom{0}} & \rawquant{1}{0.000} & \rawquant{3}{0.371} & \rawquant{3.5}{0.542} & \rawquant{2}{2.117} & \rawquant{4.5}{\phantom{0}} \\
\bf Ref & la ciccia is a moderate price restaurant in bernal heights                    \\
\bf\Bstrut Out & la ciccia, is in the bernal heights area with a moderate price range.  \\\hdashline[0.5pt/2pt]

\Tstrut\bf MR & inform(name=`intercontinental san francisco',price\_range=`pricey')
& \rawquant{2}{\phantom{0}} & \rawquant{4.5}{0.707} & \rawquant{3}{0.433} & \rawquant{5.5}{0.875} & \rawquant{2}{2.318} & \rawquant{5}{\phantom{0}} \\
\bf Ref & sure, the intercontinental san francisco is in the pricey range. \\
\bf Out & the intercontinental san francisco is in the pricey price range. \\\hline
\end{tabular}
\end{center}
\caption{Examples from our dataset. \emph{Ref} = human reference (one selected), \emph{Out} = system output to be rated. \emph{H} = median human rating, \emph{B} = BLEU, \emph{M} = METEOR, \emph{R} = ROUGE, \emph{C} = CIDEr, \emph{\ref{li:Ftonly}} = rating given by our system in the \ref{li:Ftonly} configuration. The metrics are shown with normalised and rounded values (see Section~\ref{sec:eval-measures}) on top and the original, raw values underneath (in italics). The top example is rated low by all metrics, our system is more accurate. The bottom one is rated low by humans but high by some metrics and our system.}\label{fig:data-examples}
\end{figure*}

Using the CrowdFlower crowdsourcing platform,\footnote{\url{http://www.crowdflower.com}} we collected a dataset of human rankings for outputs of three recent data-driven NLG systems as provided to us by the systems' authors; see \cite{novikova:2017} for more details.
The following systems are included in our set:%\OD{maybe cite our metrics paper?}
\begin{itemize}[itemsep=0pt,topsep=0pt]
\item LOLS \cite{lampouras_imitation_2016}, which is based on imitation learning,
\item RNNLG \cite{wen_semantically_2015}, a RNN-based system,
\item TGen \cite{dusek_training_2015}, a system using perceptron-guided incremental tree generation.
\end{itemize}
Their outputs are on the test parts of the following datasets (see Table~\ref{tab:data-stats}):
\begin{itemize}[itemsep=0pt,topsep=0pt]
\item BAGEL \cite{mairesse_phrase-based_2010} -- 404 short text segments (1--2 sentences) informing about restaurants,
\item SFRest \cite{wen_semantically_2015} -- ca.~5,000 segments from the restaurant information domain (including questions, confirmations, greetings, etc.),
\item SFHot \cite{wen_semantically_2015} -- a set from the hotel domain similar in size and contents to SFRest.
\end{itemize}
%The LOLS system outputs were provided for all datasets, TGen outputs only for BAGEL, and RNNLG outputs only for SFRest and SFHot.

During the crowdsourcing evaluation of the outputs, crowd workers were given two random system outputs along with the source MRs and were asked to evaluate the absolute overall quality of both outputs on a 1--6 Likert scale (see Figure~\ref{fig:data-examples}).
We collected at least three ratings for each
system output; this resulted in more ratings for the same sentence if two systems' outputs were identical. 
The ratings show a moderate inter-rater agreement of  0.45 ($p<0.001$) intra-class correlation coefficient \cite{landis1977measurement} across all three datasets.
We computed medians of the three (or more) ratings in our experiments to ensure more consistent ratings, which resulted in .5 ratings for some examples. We keep this granularity throughout our experiments.

We use our data in a 5-fold cross-validation setting (three training, one development, and one test part in each fold). 
We also test our model on a subset of ratings for a particular NLG system or dataset in order to assess its cross-system and cross-domain performance (see Section~\ref{sec:results}).

\subsection{Data Preprocessing}\label{sec:preprocessing}

The source MRs in our data are variants of the dialogue acts (DA) formalism \cite{young_hidden_2010} -- a shallow domain-specific MR, consisting of the main DA type (\emph{hello}, \emph{inform}, \emph{request}) and an optional set of slots (attributes, such as \emph{food} or \emph{location}) and values (e.g.\ \emph{Chinese} for food or \emph{city centre} for location).
DAs are converted into sequences for our system as a list of triplets ``DA type -- slot -- value'', where DA type may be repeated multiple times and/or special null tokens are used if slots or values are not present (see Figure~\ref{fig:model}).
The system outputs are tokenised and lowercased for our purposes.

We use delexicalisation to prevent data sparsity, following \cite{mairesse_phrase-based_2010,henderson_robust_2014,wen_semantically_2015},
where values of most DA slots (except unknown and binary \emph{yes/no} values) are replaced in both the source MRs and the system outputs by slot placeholders -- e.g.\ \emph{Chinese} is replaced by \emph{X-food} (cf.\ also Figure~\ref{fig:model}).\footnote{Note that only values matching the source MR are delexicalised in the system outputs -- if the outputs contain values not present in the source MR, they are kept intact in the model input.}

\subsection{Synthesising Additional Training Data}\label{sec:synth-data}

\begin{table}
\begin{center}
\begin{tabular}{lcccccc}\hline
\bf Setup       & \bf Instances\Tstrut\Bstrut\\\hline
\ref{li:base}   & \phantom{0}1,476 \\
\ref{li:fs}     & \phantom{0}3,937 \\
\ref{li:fts}    & 13,442 \\
\ref{li:Ftonly} & 45,137 \\
\ref{li:Ftrain} & 57,372 \\
\ref{li:Fall}   & 80,522 \\\hline
\end{tabular}
\end{center}
\caption{Training data size comparison for different data augmentation procedures (varies slightly in different cross-validation folds due to rounding).}\label{tab:data-size}
\vspace{-0.3cm}
\end{table}

Following prior works in grammatical error correction \cite{rozovskaya_ui_2012,felice_generating_2014,xie_neural_2016}, we synthesise additional training instances by introducing artificial errors:
Given a training instance (source MR, system output, and human rating), we generate a number of errors in the system output and lower the human rating accordingly.
We use a set of basic heuristics mimicking some of the observed system behaviour to introduce errors into the system outputs:\footnote{To ensure that the errors truly are detrimental to the system output quality, our rules prioritise content words, i.e.\ they do not change articles or punctuation if other words are present. The rules never remove the last word left in the system output.}
\begin{enumerate}[itemsep=0pt,topsep=0pt]
\item removing a word,
\item duplicating a word at its current position,
\item duplicating a word at a random position,% in the sentence
\item adding a random word from a dictionary learned from all training system outputs,
\item replacing a word with a random word from the dictionary.
\end{enumerate}
We lower the original Likert scale rating of the instance by 1 for each generated error. If the original rating was 5.5 or 6, the rating is lowered to 4 with the first introduced error and by 1 for each additional error.

We also experiment with using additional natural language sentences where human ratings are not available -- we use human-authored references from the original training datasets and assume that these would receive the maximum rating of 6.
We introduce artificial errors into the human references in exactly the same way as with training system outputs.

\subsection{Model Training Parameters}\label{sec:training-params}

We set the network parameters based on several experiments performed on the development set of one of the cross-validation folds (see Section~\ref{sec:dataset}).\footnote{We use embedding size 300, learning rate 0.0001, dropout probability 0.5, and two fully connected $\tanh$ layers ($k=2$).}

We train the network for 500 passes over the training data, checking Pearson and Spearman correlations on the validation set after each pass (with equal importance). We keep the configuration that yielded the best correlations overall.
For setups using synthetic training data (see Section~\ref{sec:synth-data}), we first perform 20 passes over all data including synthetic, keeping the best parameters, and then proceed with 500 passes over the original data.
To compensate for the effects of random network initialisation, all our results are averaged over five runs with different initial random seeds following \citet{wen_semantically_2015}.

\subsection{Evaluation Measures}\label{sec:eval-measures}

Following practices from MT Quality estimation \cite{bojar_findings_2016},\footnote{See also the currently ongoing WMT`17 Quality Estimation shared task at \url{http://www.statmt.org/wmt17/quality-estimation-task.html}.} we use Pearson's correlation of the predicted output quality with median human ratings as our primary evaluation metric. Mean absolute error (MAE), root mean squared error (RMSE), and Spearman's rank correlation are used as additional metrics.
%\footnote{We measure Spearman's correlation even though our system is not trained to rank a list of outputs, but predict their absolute quality. \OD{TODO does this actually make sense at all?}}
%\OD{Use prediction accuracy (percentage of spot-on ratings) as well?}

We compare our results to some of the common word-overlap metrics -- BLEU \cite{papineni_bleu_2002}, METEOR \cite{lavie_meteor:_2007}, ROUGE-L \cite{lin_rouge:_2004}, and CIDEr \cite{vedantam_cider:_2015} -- normalised into the 1--6 range of the predicted human ratings and further rounded to 0.5 steps.\footnote{We used the Microsoft COCO Captions evaluation script to obtain the metrics scores \cite{chen_microsoft_2015}. Trials with non-quantised metrics yielded very similar correlations.}
In addition, we also show the MAE/RMSE values for a trivial constant baseline that always predicts the overall average human rating (4.5).

\section{Results}\label{sec:results}

\begin{table*}
\begin{center}
\begin{tabular}{lcccc}\hline
\bf Setup                                                       & \bf Pearson & \bf Spearman & \bf MAE & \bf RMSE \\\hline
Constant                                                        & -      & -      & 1.013 & 1.233 \\
BLEU*                                                           & 0.074  & 0.061  & 2.264 & 2.731 \\
METEOR*                                                         & 0.095  & 0.099  & 1.820 & 2.129 \\
ROUGE-L*                                                        & 0.079  & 0.072  & 1.312 & 1.674 \\
CIDEr*                                                          & 0.061  & 0.058  & 2.606 & 2.935 \\\hdashline[0.5pt/2pt]
%
% original submission
%\ref{li:base}: Base system                                      & 0.259  & 0.255  & 0.991 & 1.327 \\ % 714
%\ref{li:fs}: + errors generated in training system outputs      & \bf 0.283  & 0.272  & 0.960 & 1.274 \\ % 715
%\ref{li:fts}: + training references, with generated errors      & 0.287  & 0.268  & 0.970 & 1.291 \\ % 769
%\ref{li:Ftonly}: + systems training data, with generated errors & \bf 0.355  & \bf 0.310  & 0.923 & 1.238 \\\hdashline[0.5pt/2pt] % 770
%noref_fhs & 0.323  & 0.287  & 0.937/0.798 & 0.178  \\ % 716
%\ref{li:Ftrain}: + test references, with generated errors*      & \bf 0.347  & 0.285  & 0.944 & 1.266 \\ % 717
%\ref{li:Fall}: + complete datasets, with generated errors*      & \bf 0.373  & \bf 0.297  & 0.894 & 1.213 \\\hline % 718 (+devel +once more test refs, but not relex settings)
%
% data bugfix (proper delex, random inits)
\ref{li:base}: Base system                                      & 0.273 & 0.260 & 0.948 & 1.258 \\ %818
\ref{li:fs}: + errors generated in training system outputs      & 0.283 & 0.268 & 0.948 & 1.273 \\ %819
\ref{li:fts}: + training references, with generated errors      & 0.278 & 0.261 & 0.930 & 1.257 \\ %820
\ref{li:Ftonly}: + systems training data, with generated errors & \bf 0.330 & 0.274 & 0.914 & 1.226 \\\hdashline[0.5pt/2pt] %821
\ref{li:Ftrain}: + test references, with generated errors*      & \bf 0.331 & 0.265 & 0.937 & 1.245 \\ %822
\ref{li:Fall}: + complete datasets, with generated errors*      & \bf 0.354 & \bf 0.287 & 0.909 & 1.208 \\\hline %823
\end{tabular}
\end{center}
\caption{Results using cross-validation over the whole dataset. Setups marked with ``*'' use human references for the test instances. All setups \ref{li:base}--\ref{li:Fall} produce significantly better correlations than all metrics ($p<0.01$). Significant improvements in correlation ($p<0.05$) over \ref{li:base} are marked in bold.}\label{tab:results-cv}
\end{table*}

We test the following configurations that only differ in the amount and nature of synthetic data used (see Section~\ref{sec:synth-data} and Table~\ref{tab:data-size}):
\begin{enumerate}[label=S\arabic*,itemsep=0pt,topsep=0pt]
\item \label{li:base} Base system variant, with no synthetic data.
\item \label{li:fs} Adding synthetic data -- introducing artificial errors into system outputs from the training portion of our dataset (no additional human references are used).
\item \label{li:fts} Same as previous, but with additional human references from the training portion of our dataset (including artificial errors; see Section~\ref{sec:synth-data}).\footnote{As mentioned in Section~\ref{sec:dataset}, our dataset only comprises the test sets of the source NLG datasets, i.e.\ the additional human references in \ref{li:fts} represent a portion of the source test sets. The difference to \ref{li:Ftonly} is the amount of the additional data (see Table~\ref{tab:data-size}).\label{fn:fts-vs-ftonly}}
\item \label{li:Ftonly} As previous, but with additional human references from the training parts of the respective source NLG datasets (including artificial errors), i.e.\ references on which the original NLG systems were trained.\textsuperscript{\ref{fn:fts-vs-ftonly}}
\item \label{li:Ftrain} As previous, but also includes additional human references from the test portion of our dataset (including artificial errors).\footnote{Note that the model still does not have any access at training time to test NLG system outputs or their true ratings.}
\item \label{li:Fall} As previous, but also includes development parts of the source NLG datasets (including artificial errors).
\end{enumerate}
Synthetic data are never created from system outputs in the test part of our dataset.
Note that~\ref{li:base} and~\ref{li:fs} only use the original system outputs and their ratings, with no additional human references. \ref{li:fts}~and~\ref{li:Ftonly} use additional human references (i.e.\ more in-domain data), but do not use human references for the instances on which the system is tested. \ref{li:Ftrain}~and~\ref{li:Fall} also use human references for test MRs, even if not directly, and are thus not strictly referenceless.

\subsection{Results using the whole dataset}\label{sec:cv}

The correlations and error values we obtained over the whole data in a cross-validation setup are shown in Table~\ref{tab:results-cv}. 
The correlations only stay moderate for all system variants.
On the other hand, we can see that even the base setup (\ref{li:base}) trained using less than 2,000 examples performs better than all the word-overlap metrics in terms of all evaluation measures. Improvements in both Pearson and Spearman correlations are significant according to the Williams test \cite{williams_regression_1959,kilickaya_re-evaluating_2017} ($p<0.01$). When comparing the base setup against the constant baseline, MAE is lower but RMSE is slightly higher, which suggests that our system does better on average but is prone to occasional large errors.

The results also show that the performance can be improved considerably by adding synthetic data, especially after more than tripling the training data in \ref{li:Ftonly} (Pearson correlation improvements are statistically significant in terms of the Williams test, $p<0.01$). Using additional human references for the test data seems to be helping further in \ref{li:Fall} (the difference in Pearson correlation is statistically significant, $p<0.01$):
The additional references apparently provide more information even though the SFHot and SFRest datasets have similar MRs (identical when delexicalised) in training and test data \cite{lampouras_imitation_2016}.\footnote{Note that unlike in NLG systems training on SFHot and SFRest, the identical MRs in training and test data do not allow our system to only memorize the training data as the NLG outputs to be rated are distinct. However, the situation is not 100\% referenceless as the system may have been exposed to other NLG outputs for the same MR. Our preliminary experiments suggest that our systems can also handle lexicalised data well, without any modification (Pearson correlation 0.264--0.359 for \ref{li:base}--\ref{li:Fall}).}
The setups using larger synthetic data further improve MAE and RMSE: \ref{li:Ftonly} and \ref{li:Fall} increase the margin against the constant baseline up to ca.~0.1 in terms of MAE, and both are able to surpass the constant baseline in terms of RMSE.

\subsection{Cross-domain and Cross-System Training}\label{sec:cross-domain}

%\OD{The measurements should be recomputed so that the test set is always exactly the same (do this next time)!}

\begin{table*}[tb]
\begin{center}\small
\begin{tabular}{l cccc cccc cccc}\hline
& \multicolumn{4}{c}{\ref{it:xO}: small in-domain data only} & \multicolumn{4}{c}{\ref{it:xT}: out-of-domain data only} & \multicolumn{4}{c}{\ref{it:xA}: out-of-dom.\ + small in-dom.} \\ 
& \bf Pear & \bf Spea & \bf MAE & \bf RMSE & \bf Pear & \bf Spea & \bf MAE & \bf RMSE & \bf Pear & \bf Spea & \bf MAE & \bf RMSE \\\hline
% original submision
%Constant              & -     & -     & 0.998 & 1.229 & -     & -     & 0.990 & 1.217 & -     & -     & 0.998 & 1.229 \\
%BLEU*                 & 0.045 & 0.030 & 2.241 & 2.706 & 0.061 & 0.047 & 2.214 & 2.692 & 0.045 & 0.030 & 2.241 & 2.706 \\
%METEOR*\hspace{-3mm}  & 0.062 & 0.058 & 1.741 & 2.052 & 0.089 & 0.084 & 1.727 & 2.040 & 0.062 & 0.058 & 1.741 & 2.052 \\
%ROUGE-L*\hspace{-3mm} & 0.043 & 0.021 & 1.282 & 1.637 & 0.076 & 0.061 & 1.248 & 1.612 & 0.043 & 0.021 & 1.282 & 1.637 \\
%CIDEr*                & 0.029 & 0.016 & 2.612 & 2.946 & 0.062 & 0.060 & 2.643 & 2.960 & 0.029 & 0.016 & 2.612 & 2.946 \\\hdashline[0.5pt/2pt]
% after using the same test set for everything
Constant              & -     & -     & 0.994 & 1.224 & -     & -     & 0.994 & 1.224 & -     & -     & 0.994 & 1.224 \\
BLEU*                 & 0.033 & 0.016 & 2.235 & 2.710 & 0.033 & 0.016 & 2.235 & 2.710 & 0.033 & 0.016 & 2.235 & 2.710 \\
METEOR*               & 0.076 & 0.074 & 1.719 & 2.034 & 0.076 & 0.074 & 1.719 & 2.034 & 0.076 & 0.074 & 1.719 & 2.034 \\
ROUGE-L*              & 0.064 & 0.049 & 1.255 & 1.620 & 0.064 & 0.049 & 1.255 & 1.620 & 0.064 & 0.049 & 1.255 & 1.620 \\
CIDEr*                & 0.048 & 0.043 & 2.590 & 2.921 & 0.048 & 0.043 & 2.590 & 2.921 & 0.048 & 0.043 & 2.590 & 2.921 \\\hdashline[0.5pt/2pt]
% original submission
%\ref{li:base}         & 0.168 & 0.172 & 1.130 & 1.459 & 0.145 & 0.154 & 1.017 & 1.360 & 0.184 & 0.185 & 1.024 & 1.340 \\
%\ref{li:fs}           & 0.150 & 0.162 & 1.210 & 1.558 & 0.164 & \bf 0.176 & 1.026 & 1.355 & \bf \uline{0.220} & \bf \uline{0.226} & 0.989 & 1.303 \\
%\ref{li:fts}          & 0.120 & 0.124 & 1.093 & 1.362 & 0.158 & 0.178 & 1.100 & 1.372 & \bf \uline{0.209} & \bf \uline{0.216} & 1.019 & 1.351 \\
%\ref{li:Ftonly}       & \bf 0.188 & 0.160 & 1.048 & 1.383 & \bf 0.192 & 0.146 & 1.055 & 1.407 & \bf \uline{0.265} & \bf \uline{0.232} & 1.001 & 1.337 \\\hdashline[0.5pt/2pt]
%\ref{li:Ftrain}*      & \bf 0.243 & \bf 0.212 & 1.030 & 1.359 & \bf 0.195 & 0.144 & 1.141 & 1.512 & \bf 0.253 & \bf 0.229 & 0.999 & 1.324 \\
%\ref{li:Fall}*        & \bf 0.242 & \bf 0.206 & 1.054 & 1.386 & 0.140 & 0.106 & 1.152 & 1.522 & \bf \uline{0.282} & \bf \uline{0.246} & 1.000 & 1.337 \\\hline
% fixed
\ref{li:base}         & 0.147 & 0.136 & 1.086 & 1.416 & 0.162 & 0.152 & 0.985 & 1.281 & 0.170 & \uline{0.166} & 1.003 & 1.315 \\ %778 802 784
\ref{li:fs}           & \bf 0.196 & \bf 0.176 & 1.059 & 1.364 & \bf 0.197 & \bf 0.189 & 1.003 & 1.311 & \bf 0.219 & \bf \uline{0.218} & 0.988 & 1.296 \\ %779 803 785
\ref{li:fts}          & \bf 0.176 & \bf 0.163 & 1.093 & 1.420 & 0.147 & 0.134 & 1.037 & 1.366 & \bf \uline{0.219} & \bf \uline{0.216} & 0.979 & 1.302 \\ % 848 850 852
\ref{li:Ftonly}       & \bf 0.264 & \bf 0.218 & 0.983 & 1.307 & 0.162 & 0.138 & 1.084 & 1.448 & \bf 0.247 & \bf 0.211 & 0.983 & 1.306 \\\hdashline[0.5pt/2pt] % 849 851 853
\ref{li:Ftrain}*      & \bf 0.280 & \bf 0.221 & 1.009 & 1.341 & 0.173 & 0.145 & 1.077 & 1.438 & \bf 0.210 & 0.162 & 1.095 & 1.442 \\ %782 806 788
\ref{li:Fall}*        & \bf 0.271 & \bf 0.202 & 0.991 & 1.331 & 0.188 & 0.178 & 1.037 & 1.392 & \bf 0.224 & \bf 0.210 & 1.002 & 1.339 \\\hline %783 807 789
\end{tabular}
\end{center}
\caption{Cross-domain evaluation results. Setups marked with ``*'' use human references of test instances. All setups produce significantly better correlations than all metrics ($p<0.01$). Significant improvements in correlation ($p<0.05$) over the corresponding \ref{li:base} are marked in bold, significant improvements over the corresponding \ref{it:xO} are underlined.}\label{tab:cross-domain}
\end{table*}

\begin{table*}[tb]
\begin{center}\small
\begin{tabular}{l cccc cccc cccc}\hline
& \multicolumn{4}{c}{\ref{it:xO}: small in-system data only} & \multicolumn{4}{c}{\ref{it:xT}: out-of-system data only} & \multicolumn{4}{c}{\ref{it:xA}: out-of-sys.\ + small in-sys.} \\ 
& \bf Pear & \bf Spea & \bf MAE & \bf RMSE & \bf Pear & \bf Spea & \bf MAE & \bf RMSE & \bf Pear & \bf Spea & \bf MAE & \bf RMSE \\\hline
% original submission
%Constant              & -     & -     & 1.060 & 1.301 & -     & -     & 1.063 & 1.297 & -     & -     & 1.060 & 1.301 \\
%BLEU*                 & 0.065 & 0.029 & 2.508 & 2.973 & 0.084 & 0.049 & 2.490 & 2.956 & 0.065 & 0.029 & 2.508 & 2.973 \\
%METEOR*\hspace{-3mm}  & 0.114 & 0.098 & 1.942 & 2.250 & 0.132 & 0.117 & 1.922 & 2.234 & 0.114 & 0.098 & 1.942 & 2.250 \\
%ROUGE-L*\hspace{-3mm} & 0.046 & 0.022 & 1.453 & 1.808 & 0.063 & 0.047 & 1.433 & 1.790 & 0.046 & 0.022 & 1.453 & 1.808 \\
%CIDEr*                & 0.106 & 0.083 & 2.820 & 3.131 & 0.123 & 0.105 & 2.802 & 3.114 & 0.106 & 0.083 & 2.820 & 3.131 \\\hdashline[0.5pt/2pt]
% fixed
Constant              & -     & -     & 1.060 & 1.301 & -     & -     & 1.060 & 1.301 & -     & -     & 1.060 & 1.301 \\
BLEU*                 & 0.079 & 0.043 & 2.514 & 2.971 & 0.079 & 0.043 & 2.514 & 2.971 & 0.079 & 0.043 & 2.514 & 2.971 \\
METEOR*               & 0.141 & 0.122 & 1.929 & 2.238 & 0.141 & 0.122 & 1.929 & 2.238 & 0.141 & 0.122 & 1.929 & 2.238 \\
ROUGE-L*              & 0.064 & 0.048 & 1.449 & 1.802 & 0.064 & 0.048 & 1.449 & 1.802 & 0.064 & 0.048 & 1.449 & 1.802 \\
CIDEr*                & 0.127 & 0.106 & 2.801 & 3.112 & 0.127 & 0.106 & 2.801 & 3.112 & 0.127 & 0.106 & 2.801 & 3.112 \\\hdashline[0.5pt/2pt]
% original submission
%\ref{li:base}         & 0.276 & 0.269 & 1.107 & 1.458 & \it 0.054 & \it 0.072 & 1.117 & 1.421 & 0.205 & 0.232 & 1.082 & 1.416 \\
%\ref{li:fs}           & \bf 0.331 & \bf 0.329 & 1.020 & 1.349 & \bf\it 0.116 & \bf\it 0.116 & 1.118 & 1.461 & 0.202 & 0.210 & 1.288 & 1.592 \\
%\ref{li:fts}          & \bf 0.295 & \bf 0.280 & 1.031 & 1.375 & \bf 0.148 & \bf 0.163 & 1.076 & 1.409 & \bf 0.240 & \bf 0.257 & 1.090 & 1.451 \\
%\ref{li:Ftonly}       & \bf 0.336 & \bf 0.315 & 1.043 & 1.407 & \bf 0.276 & \bf 0.256 & 1.006 & 1.326 & \bf 0.343 & \bf 0.320 & 0.981 & 1.314 \\\hdashline[0.5pt/2pt]
%\ref{li:Ftrain}*      & \bf 0.371 & \bf 0.351 & 0.993 & 1.315 & \bf 0.265 & \bf 0.235 & 1.004 & 1.312 & \bf 0.348 & \bf 0.328 & 1.018 & 1.322 \\
%\ref{li:Fall}*        & \bf 0.388 & \bf 0.351 & 0.982 & 1.316 & \bf 0.234 & \bf 0.214 & 1.028 & 1.364 & \bf 0.334 & \bf 0.318 & 1.009 & 1.336 \\\hline
% fixed
\ref{li:base}         & \it 0.341 & 0.334 & 1.054 & 1.405 & \it 0.097 & \it 0.117 & 1.052 & 1.336 & 0.174 & 0.179 & 1.114 & 1.455 \\ %790 808 796
\ref{li:fs}           & \bf 0.358 & 0.345 & 1.007 & 1.342 & \it 0.115 & \it 0.119 & 1.057 & 1.355 & \bf 0.203 & \bf 0.222 & 1.253 & 1.613 \\ %791 809 797
\ref{li:fts}          & \bf 0.378 & \bf 0.365 & 0.971 & 1.326 & \it 0.112 & \it 0.094 & 1.059 & 1.387 & \bf \uline{0.404} & \bf 0.377 & 0.968 & 1.277 \\ % 854 856 858
\ref{li:Ftonly}       & \bf 0.390 & \bf 0.360 & 0.981 & 1.311 & \bf 0.247 & \bf 0.189 & 1.011 & 1.338 & \bf 0.370 & \bf 0.346 & 0.997 & 1.312 \\\hdashline[0.5pt/2pt] % 855 857 859
\ref{li:Ftrain}*      & \bf 0.398 & \bf 0.364 & 1.043 & 1.393 & \bf 0.229 & \bf 0.174 & 1.025 & 1.328 & \bf 0.386 & \bf 0.356 & 0.975 & 1.301 \\ %794 812 800
\ref{li:Fall}*        & \bf 0.390 & 0.353 & 1.036 & 1.389 & \bf 0.332 & \bf 0.262 & 0.969 & 1.280 & \bf 0.374 & \bf 0.330 & 0.979 & 1.298 \\\hline %795 813 801
\end{tabular}
\end{center}
\caption{Cross-system evaluation results. Setups marked with ``*'' use human references of test instances. Setups that do not produce significantly better correlations than all metrics ($p<0.05$) are marked in italics. Significant improvements in correlation ($p<0.05$) over the corresponding \ref{li:base} are marked in bold, significant improvements over the corresponding \ref{it:xO} are underlined.}\label{tab:cross-system}
\end{table*}

Next, we test how well our approach generalises to new systems and datasets and how much in-set data (same domain/system) is needed to obtain reasonable results.
We use the SFHot data as our test domain and LOLS as our test system, and we treat the rest as out-of-set.
We test three different configurations:
\begin{enumerate}[label=C\arabic*,itemsep=0pt,topsep=0pt]
\item \label{it:xO} Training exclusively using a small amount of in-set data (200 instances, 100 reserved for validation), testing on the rest of the in-set.
\item \label{it:xT} Training and validating exclusively on out-of-set data, testing on the same part of the in-set as in \ref{it:xO} and \ref{it:xA}.
\item \label{it:xA} Training on the out-of-set data with a small amount of in-set data (200 instances, 100 reserved for validation), testing on the rest of the in-set.
\end{enumerate}
The results are shown in Tables~\ref{tab:cross-domain} and~\ref{tab:cross-system}, respectively.

The correlations of \ref{it:xT} suggest that while our network can generalise across systems to some extent (if data fabrication is used), it does not generalise well across domains without using in-domain training data.
\ref{it:xO}~and \ref{it:xA}~configuration results demonstrate that even small amounts of in-set data help noticeably.
However, if in-set data is used, additional out-of-set data does not improve the results in most cases (\ref{it:xA} is mostly not significantly better than the corresponding \ref{it:xO}).

Except for a few cross-system \ref{it:xT} configurations with low amounts of synthetic data, all systems perform better than word-overlap metrics. However, most setups are not able to improve over the constant baseline in terms of MAE and RMSE.

\section{Related Work}\label{sec:related}

This work is the first NLG QE system to our knowledge; the most  related work in NLG is probably the system by \citet{dethlefs2014cluster}, which reranks NLG outputs by estimating their properties (such as colloquialism or politeness) using various regression models.
However, our work is also related to QE research in other areas, such as MT \cite{specia_machine_2010}, dialogue systems \cite{lowe_towards_2017} or grammatical error correction  \cite{napoles_theres_2016}.
QE is especially well researched for MT, where regular QE shared tasks are organised \cite{callison-burch_findings_2012,bojar_findings_2013,bojar_findings_2014,bojar_findings_2015,bojar_findings_2016}.

Many of the past MT QE systems participating in the shared tasks are based on Support Vector Regression \cite{specia_multi-level_2015,bojar_findings_2014,bojar_findings_2015}. Only in the past year, NN-based solutions started to emerge. 
\citet{patel_translation_2016} present a system based on RNN language models, which focuses on predicting MT quality on the word level. 
\citet{kim_recurrent_2016} estimate segment-level MT output quality using a bidirectional RNN over both source and output sentences combined with a logistic prediction layer. 
They pretrain their RNN on large MT training data.

Last year's MT QE shared task systems achieve Pearson correlations between 0.4--0.5, which is slightly higher than our best results. However, the results are not directly comparable: First, we predict a Likert-scale assessment instead of the number of required post-edits. Second, NLG datasets are considerably smaller than corpora available for MT. Third, we believe that QE for NLG is harder due to the reasons outlined in Section~\ref{sec:intro}.

%\OD{TODO citations suggested by VR: (Callison-burch  et  al.,  2012;  Bojar  et  al.,  2013 https://arxiv.org/pdf/1606.09600.pdf
%Last year's edition featured mostly RNNs. http://www.statmt.org/wmt16/pdf/W16-2301.pdf}

\section{Conclusions and Future Work}\label{sec:concl}

We presented the first system for referenceless quality estimation of natural language generation outputs. 
All code and data used here is available online at:
\begin{center}
\url{https://github.com/tuetschek/ratpred}
\end{center}

In an evaluation spanning outputs of three different NLG systems and three datasets,
our system significantly outperformed four commonly used reference-based metrics.
It also improved over a constant baseline, which always predicts the mean human rating, in terms of MAE and RMSE. The smaller RMSE improvement suggests that our system is prone to occasional large errors.
We have shown that generating additional training data, e.g.\ by using NLG training datasets and synthetic errors, significantly improves the system performance.
While our system can generalise to unseen NLG systems in the same domain to some extent, its cross-domain generalisation capability is poor. 
However, very small amounts of in-domain/in-system data improve performance notably.

In future work, we will explore improvements to our error synthesising methods as well as changes to our network architecture (bidirectional RNNs or convolutional NNs).
We also plan to focus on relative ranking of different NLG outputs for the same source MR or predicting the number of post-edits required.
We intend to use data collected within the ongoing E2E NLG Challenge \cite{novikova_e2e_2017}, which promises greater diversity than current datasets.

%\subsection{Software and Data}
% provide in supplementary material or anonymous URL

%\section*{Acknowledgements} 

%\footnotesize
\section*{Acknowledgements}
%\vspace{-0.2cm}
The authors would like to thank Lucia Specia and the two anonymous reviewers for their helpful comments.
This research received funding from the EPSRC projects  DILiGENt (EP/M005429/1) and  MaDrIgAL (EP/N017536/1). The Titan Xp used for this research was donated by the NVIDIA Corporation.
%\normalsize

\bibliography{references}

\begin{thebibliography}{41}
\providecommand{\natexlab}[1]{#1}
\providecommand{\url}[1]{\texttt{#1}}
\expandafter\ifx\csname urlstyle\endcsname\relax
  \providecommand{\doi}[1]{doi: #1}\else
  \providecommand{\doi}{doi: \begingroup \urlstyle{rm}\Url}\fi

\bibitem[Bengio et~al.(2003)Bengio, Ducharme, Vincent, and
  Jauvin]{bengio_neural_2003}
Bengio, Yoshua, Ducharme, Réjean, Vincent, Pascal, and Jauvin, Christian.
\newblock A {Neural} {Probabilistic} {Language} {Model}.
\newblock \emph{Journal of Machine Learning Research}, 3:\penalty0 1137--1155,
  February 2003.

\bibitem[Bojar et~al.(2016)Bojar, Chatterjee, Federmann, Graham, Haddow, Huck,
  Yepes, Koehn, Logacheva, Monz, and {others}]{bojar_findings_2016}
Bojar, Ondrej, Chatterjee, Rajen, Federmann, Christian, Graham, Yvette, Haddow,
  Barry, Huck, Matthias, Yepes, Antonio~Jimeno, Koehn, Philipp, Logacheva,
  Varvara, Monz, Christof, and {others}.
\newblock Findings of the 2016 conference on machine translation ({WMT}16).
\newblock In \emph{Proceedings of the {First} {Conference} on {Machine}
  {Translation} ({WMT})}, pp.\  131--198, Berlin, Germany, 2016.

\bibitem[Bojar et~al.(2013)Bojar, Buck, Callison-Burch, Federmann, Haddow,
  Koehn, Monz, Post, Soricut, and Specia]{bojar_findings_2013}
Bojar, Ond\v{r}ej, Buck, Christian, Callison-Burch, Chris, Federmann,
  Christian, Haddow, Barry, Koehn, Philipp, Monz, Christof, Post, Matt,
  Soricut, Radu, and Specia, Lucia.
\newblock Findings of the 2013 {Workshop} on {Statistical} {Machine}
  {Translation}.
\newblock In \emph{Proceedings of the {Eighth} {Workshop} on {Statistical}
  {Machine} {Translation}}, pp.\  1--44, Sofia, Bulgaria, 2013.

\bibitem[Bojar et~al.(2014)Bojar, Buck, Federmann, Haddow, Koehn, Leveling,
  Monz, Pecina, Post, Saint-Amand, Soricut, Specia, and
  Tamchyna]{bojar_findings_2014}
Bojar, Ond\v{r}ej, Buck, Christian, Federmann, Christian, Haddow, Barry, Koehn,
  Philipp, Leveling, Johannes, Monz, Christof, Pecina, Pavel, Post, Matt,
  Saint-Amand, Herve, Soricut, Radu, Specia, Lucia, and Tamchyna, Ale\v{s}.
\newblock Findings of the 2014 {Workshop} on {Statistical} {Machine}
  {Translation}.
\newblock In \emph{Proceedings of the {Ninth} {Workshop} on {Statistical}
  {Machine} {Translation}}, pp.\  12--58, Baltimore, MD, USA, 2014.

\bibitem[Bojar et~al.(2015)Bojar, Chatterjee, Federmann, Haddow, Huck, Hokamp,
  Koehn, Logacheva, Monz, Negri, Post, Scarton, Specia, and
  Turchi]{bojar_findings_2015}
Bojar, Ond\v{r}ej, Chatterjee, Rajen, Federmann, Christian, Haddow, Barry,
  Huck, Matthias, Hokamp, Chris, Koehn, Philipp, Logacheva, Varvara, Monz,
  Christof, Negri, Matteo, Post, Matt, Scarton, Carolina, Specia, Lucia, and
  Turchi, Marco.
\newblock Findings of the 2015 {Workshop} on {Statistical} {Machine}
  {Translation}.
\newblock In \emph{Proceedings of the {Tenth} {Workshop} on {Statistical}
  {Machine} {Translation}}, pp.\  1--46, Lisbon, Portugal, 2015.

\bibitem[Callison-Burch et~al.(2012)Callison-Burch, Koehn, Monz, Post, Soricut,
  and Specia]{callison-burch_findings_2012}
Callison-Burch, Chris, Koehn, Philipp, Monz, Christof, Post, Matt, Soricut,
  Radu, and Specia, Lucia.
\newblock Findings of the 2012 {Workshop} on {Statistical} {Machine}
  {Translation}.
\newblock In \emph{Proceedings of the {Seventh} {Workshop} on {Statistical}
  {Machine} {Translation}}, pp.\  10--51, Montréal, Canada, 2012.

\bibitem[Cercas~Curry et~al.(2017)Cercas~Curry, Hastie, and Rieser]{curry:2017}
Cercas~Curry, Amanda, Hastie, Helen, and Rieser, Verena.
\newblock A review of evaluation techniques for social dialogue systems.
\newblock Under submission, 2017.

\bibitem[Chen \& Cherry(2014)Chen and Cherry]{chen_systematic_2014}
Chen, Boxing and Cherry, Colin.
\newblock A systematic comparison of smoothing techniques for sentence-level
  {BLEU}.
\newblock In \emph{Proceedings of the {Ninth} {Workshop} on {Statistical}
  {Machine} {Translation}}, pp.\  362--367, Baltimore, MD, USA, 2014.

\bibitem[Chen et~al.(2015)Chen, Fang, Lin, Vedantam, Gupta, Dollar, and
  Zitnick]{chen_microsoft_2015}
Chen, Xinlei, Fang, Hao, Lin, Tsung-Yi, Vedantam, Ramakrishna, Gupta, Saurabh,
  Dollar, Piotr, and Zitnick, C.~Lawrence.
\newblock Microsoft {COCO} {Captions}: {Data} {Collection} and {Evaluation}
  {Server}.
\newblock \emph{arXiv:1504.00325 [cs]}, April 2015.

\bibitem[Cho et~al.(2014)Cho, van Merrienboer, Gulcehre, Bahdanau, Bougares,
  Schwenk, and Bengio]{cho_learning_2014}
Cho, Kyunghyun, van Merrienboer, Bart, Gulcehre, Caglar, Bahdanau, Dzmitry,
  Bougares, Fethi, Schwenk, Holger, and Bengio, Yoshua.
\newblock Learning {Phrase} {Representations} using {RNN} {Encoder}-{Decoder}
  for {Statistical} {Machine} {Translation}.
\newblock In \emph{Proceedings of the 2014 {Conference} on {Empirical}
  {Methods} in {Natural} {Language} {Processing} ({EMNLP})}, pp.\  1724--1734,
  Doha, Qatar, 2014.
\newblock arXiv: 1406.1078.

\bibitem[Dethlefs et~al.(2014)Dethlefs, Cuay{\'a}huitl, Hastie, Rieser, and
  Lemon]{dethlefs2014cluster}
Dethlefs, Nina, Cuay{\'a}huitl, Heriberto, Hastie, Helen~F, Rieser, Verena, and
  Lemon, Oliver.
\newblock Cluster-based prediction of user ratings for stylistic surface
  realisation.
\newblock In \emph{Proceedings of the 14th Conference of the European Chapter
  of the Association for Computational Linguistics}, pp.\  702--711,
  Gothenburg, Sweden, 2014.

\bibitem[Du\v{s}ek \& Jur\v{c}\'i\v{c}ek(2015)Du\v{s}ek and
  Jur\v{c}\'i\v{c}ek]{dusek_training_2015}
Du\v{s}ek, Ond\v{r}ej and Jur\v{c}\'i\v{c}ek, Filip.
\newblock Training a {Natural} {Language} {Generator} {From} {Unaligned}
  {Data}.
\newblock In \emph{Proceedings of the 53rd {Annual} {Meeting} of the
  {Association} for {Computational} {Linguistics} and the 7th {International}
  {Joint} {Conference} on {Natural} {Language} {Processing}}, pp.\  451--461,
  Beijing, China, 2015.

\bibitem[Felice \& Yuan(2014)Felice and Yuan]{felice_generating_2014}
Felice, Mariano and Yuan, Zheng.
\newblock Generating artificial errors for grammatical error correction.
\newblock In \emph{Proceedings of the {Student} {Research} {Workshop} at the
  14th {Conference} of the {European} {Chapter} of the {Association} for
  {Computational} {Linguistics}}, pp.\  116--126, Gothenburg, Sweden, 2014.

\bibitem[Finch et~al.(2004)Finch, Akiba, and Sumita]{finch2004does}
Finch, Andrew~M, Akiba, Yasuhiro, and Sumita, Eiichiro.
\newblock How does automatic machine translation evaluation correlate with
  human scoring as the number of reference translations increases?
\newblock In \emph{Proceedings of the Fourth International Conference on
  Language Resources and Evaluation (LREC)}, pp.\  2019--2022, Lisbon,
  Portugal, 2004.

\bibitem[Gkatzia \& Mahamood(2015)Gkatzia and Mahamood]{gkatzia_snapshot_2015}
Gkatzia, Dimitra and Mahamood, Saad.
\newblock A {Snapshot} of {NLG} {Evaluation} {Practices} 2005 - 2014.
\newblock In \emph{Proceedings of the 15th {European} {Workshop} on {Natural}
  {Language} {Generation} ({ENLG})}, pp.\  57--60, Brighton, UK, 2015.

\bibitem[Henderson et~al.(2014)Henderson, Thomson, and
  Young]{henderson_robust_2014}
Henderson, M., Thomson, B., and Young, S.
\newblock Robust dialog state tracking using delexicalised recurrent neural
  networks and unsupervised adaptation.
\newblock In \emph{Proceedings of the 2014 {IEEE} {Spoken} {Language}
  {Technology} {Workshop} ({SLT})}, pp.\  360--365, South Lake Tahoe, NV, USA,
  2014.

\bibitem[Hinton et~al.(2012)Hinton, Srivastava, Krizhevsky, Sutskever, and
  Salakhutdinov]{hinton_improving_2012}
Hinton, Geoffrey~E., Srivastava, Nitish, Krizhevsky, Alex, Sutskever, Ilya, and
  Salakhutdinov, Ruslan~R.
\newblock Improving neural networks by preventing co-adaptation of feature
  detectors.
\newblock \emph{arXiv:1207.0580 [cs]}, July 2012.

\bibitem[Hochreiter \& Schmidhuber(1997)Hochreiter and
  Schmidhuber]{hochreiter_long_1997}
Hochreiter, Sepp and Schmidhuber, Jürgen.
\newblock Long short-term memory.
\newblock \emph{Neural computation}, 9\penalty0 (8):\penalty0 1735--1780, 1997.

\bibitem[Kilickaya et~al.(2017)Kilickaya, Erdem, Ikizler-Cinbis, and
  Erdem]{kilickaya_re-evaluating_2017}
Kilickaya, Mert, Erdem, Aykut, Ikizler-Cinbis, Nazli, and Erdem, Erkut.
\newblock Re-evaluating {Automatic} {Metrics} for {Image} {Captioning}.
\newblock In \emph{Proceedings of the 15th {Conference} of the {European}
  {Chapter} of the {Association} for {Computational} {Linguistics}: {Volume} 1,
  {Long} {Papers}}, pp.\  199--209, Valencia, Spain, 2017.
\newblock arXiv:1612.07600.

\bibitem[Kim \& Lee(2016)Kim and Lee]{kim_recurrent_2016}
Kim, Hyun and Lee, Jong-Hyeok.
\newblock Recurrent {Neural} {Network} based {Translation} {Quality}
  {Estimation}.
\newblock In \emph{Proceedings of the {First} {Conference} on {Machine}
  {Translation}, {Volume} 2: {Shared} {Task} {Papers}}, pp.\  787--792, Berlin,
  Germany, 2016.

\bibitem[Lampouras \& Vlachos(2016)Lampouras and
  Vlachos]{lampouras_imitation_2016}
Lampouras, Gerasimos and Vlachos, Andreas.
\newblock Imitation learning for language generation from unaligned data.
\newblock In \emph{Proceedings of {COLING} 2016, the 26th {International}
  {Conference} on {Computational} {Linguistics}: {Technical} {Papers}}, pp.\
  1101--1112, Osaka, Japan, 2016.

\bibitem[Landis \& Koch(1977)Landis and Koch]{landis1977measurement}
Landis, J~Richard and Koch, Gary~G.
\newblock The measurement of observer agreement for categorical data.
\newblock \emph{Biometrics}, 33\penalty0 (1):\penalty0 159--174, March 1977.

\bibitem[Lavie \& Agarwal(2007)Lavie and Agarwal]{lavie_meteor:_2007}
Lavie, Alon and Agarwal, Abhaya.
\newblock Meteor: {An} {Automatic} {Metric} for {MT} {Evaluation} with {High}
  {Levels} of {Correlation} with {Human} {Judgments}.
\newblock In \emph{Proceedings of the {Second} {Workshop} on {Statistical}
  {Machine} {Translation}}, pp.\  228--231, Prague, Czech Republic, 2007.

\bibitem[Lin(2004)]{lin_rouge:_2004}
Lin, Chin-Yew.
\newblock {ROUGE}: {A} package for automatic evaluation of summaries.
\newblock In \emph{Text summarization branches out: {Proceedings} of the
  {ACL}-04 workshop}, pp.\  74--81. Barcelona, Spain, 2004.

\bibitem[Liu et~al.(2016)Liu, Lowe, Serban, Noseworthy, Charlin, and
  Pineau]{Liu:EMNLP2016}
Liu, Chia{-}Wei, Lowe, Ryan, Serban, Iulian, Noseworthy, Michael, Charlin,
  Laurent, and Pineau, Joelle.
\newblock How {NOT} to evaluate your dialogue system: An empirical study of
  unsupervised evaluation metrics for dialogue response generation.
\newblock In \emph{Proceedings of the 2016 {Conference} on {Empirical}
  {Methods} in {Natural} {Language} {Processing}}, pp.\  2122--2132, Austin,
  TX, USA, 2016.
\newblock {arXiv:1603.08023}.

\bibitem[Lowe et~al.(2017)Lowe, Noseworthy, Serban, Angelard-Gontier, Bengio,
  and Pineau]{lowe_towards_2017}
Lowe, Ryan, Noseworthy, Michael, Serban, Iulian~V., Angelard-Gontier, Nicolas,
  Bengio, Yoshua, and Pineau, Joelle.
\newblock Towards an automatic {Turing} test: learning to evaluate dialogue
  responses.
\newblock In \emph{Proceedings of 55th {Annual} {Meeting} of the {Association}
  for {Computational} {Linguistics}}, pp.\  1116--1126, Vancouver, Canada,
  2017.

\bibitem[Mairesse et~al.(2010)Mairesse, Ga\v{s}i\'c, Jur\v{c}\'i\v{c}ek,
  Keizer, Thomson, Yu, and Young]{mairesse_phrase-based_2010}
Mairesse, F., Ga\v{s}i\'c, M., Jur\v{c}\'i\v{c}ek, F., Keizer, S., Thomson, B.,
  Yu, K., and Young, S.
\newblock Phrase-based statistical language generation using graphical models
  and active learning.
\newblock In \emph{Proceedings of the 48th {Annual} {Meeting} of the
  {Association} for {Computational} {Linguistics}}, pp.\  1552--1561, Uppsala,
  Sweden, 2010.

\bibitem[Mikolov et~al.(2013)Mikolov, Chen, Corrado, and
  Dean]{mikolov_efficient_2013}
Mikolov, Tomas, Chen, Kai, Corrado, Greg, and Dean, Jeffrey.
\newblock Efficient estimation of word representations in vector space.
\newblock \emph{arXiv:1301.3781 [cs]}, January 2013.

\bibitem[Napoles et~al.(2016)Napoles, Sakaguchi, and
  Tetreault]{napoles_theres_2016}
Napoles, Courtney, Sakaguchi, Keisuke, and Tetreault, Joel.
\newblock There's {No} {Comparison}: {Reference}-less {Evaluation} {Metrics} in
  {Grammatical} {Error} {Correction}.
\newblock In \emph{Proceedings of the 2016 {Conference} on {Empirical}
  {Methods} in {Natural} {Language} {Processing}}, pp.\  2109--2115, Austin,
  TX, USA, 2016.

\bibitem[Novikova et~al.(2017{\natexlab{a}})Novikova, Cercas~Curry, Du\v{s}ek,
  and Rieser]{novikova:2017}
Novikova, Jekaterina, Cercas~Curry, Amanda, Du\v{s}ek, Ond\v{r}ej, and Rieser,
  Verena.
\newblock Why we need new evaluation metrics for {NLG}.
\newblock In \emph{Proceedings of the 2017 Conference on Empirical Methods in
  Natural Language Processing}, Copenhagen, Denmark, 2017{\natexlab{a}}.
\newblock arXiv:1707.06875.

\bibitem[Novikova et~al.(2017{\natexlab{b}})Novikova, Du\v{s}ek, and
  Rieser]{novikova_e2e_2017}
Novikova, Jekaterina, Du\v{s}ek, Ond\v{r}ej, and Rieser, Verena.
\newblock The {E2E} dataset: New challenges for end-to-end generation.
\newblock In \emph{Proceedings of the 18th {Annual} {Meeting} of the {Special}
  {Interest} {Group} on {Discourse} and {Dialogue}}, Saarbrücken, Germany,
  2017{\natexlab{b}}.
\newblock {arXiv:1706.09254}.

\bibitem[Papineni et~al.(2002)Papineni, Roukos, Ward, and
  Zhu]{papineni_bleu_2002}
Papineni, K., Roukos, S., Ward, T., and Zhu, W.-J.
\newblock {BLEU:} a method for automatic evaluation of machine translation.
\newblock In \emph{Proceedings of the 40th annual meeting of the Association
  for Computational Linguistics}, pp.\  311--318, Pennsylvania, {PA}, {USA},
  2002.

\bibitem[Patel \& M(2016)Patel and M]{patel_translation_2016}
Patel, Raj~Nath and M, Sasikumar.
\newblock Translation {Quality} {Estimation} using {Recurrent} {Neural}
  {Network}.
\newblock In \emph{Proceedings of the {First} {Conference} on {Machine}
  {Translation}, {Volume} 2: {Shared} {Task} {Papers}}, pp.\  819--824, Berlin,
  Germany, 2016.

\bibitem[Rozovskaya et~al.(2012)Rozovskaya, Sammons, and
  Roth]{rozovskaya_ui_2012}
Rozovskaya, Alla, Sammons, Mark, and Roth, Dan.
\newblock The {UI} {System} in the {HOO} 2012 {Shared} {Task} on {Error}
  {Correction}.
\newblock In \emph{Proceedings of the {Seventh} {Workshop} on {Building}
  {Educational} {Applications} {Using} {NLP}}, pp.\  272--280, Montréal,
  Canada, 2012.

\bibitem[Specia et~al.(2010)Specia, Raj, and Turchi]{specia_machine_2010}
Specia, Lucia, Raj, Dhwaj, and Turchi, Marco.
\newblock Machine translation evaluation versus quality estimation.
\newblock \emph{Machine Translation}, 24\penalty0 (1):\penalty0 39--50, March
  2010.

\bibitem[Specia et~al.(2015)Specia, Paetzold, and
  Scarton]{specia_multi-level_2015}
Specia, Lucia, Paetzold, Gustavo, and Scarton, Carolina.
\newblock Multi-level {Translation} {Quality} {Prediction} with {QuEst}++.
\newblock In \emph{Proceedings of {ACL}-{IJCNLP} 2015 {System}
  {Demonstrations}}, pp.\  115--120, Beijing, China, 2015.

\bibitem[Vedantam et~al.(2015)Vedantam, Lawrence~Zitnick, and
  Parikh]{vedantam_cider:_2015}
Vedantam, Ramakrishna, Lawrence~Zitnick, C., and Parikh, Devi.
\newblock {CIDEr}: {Consensus}-{Based} {Image} {Description} {Evaluation}.
\newblock In \emph{Proceedings of the IEEE Conference on Computer Vision and
  Pattern Recognition}, pp.\  4566--4575, Boston, MA, USA, 2015.
\newblock arXiv:1411.5726.

\bibitem[Wen et~al.(2015)Wen, Gasic, Mrk\v{s}i\'c, Su, Vandyke, and
  Young]{wen_semantically_2015}
Wen, Tsung-Hsien, Gasic, Milica, Mrk\v{s}i\'c, Nikola, Su, Pei-Hao, Vandyke,
  David, and Young, Steve.
\newblock Semantically {Conditioned} {LSTM}-based {Natural} {Language}
  {Generation} for {Spoken} {Dialogue} {Systems}.
\newblock In \emph{Proceedings of the 2015 {Conference} on {Empirical}
  {Methods} in {Natural} {Language} {Processing}}, pp.\  1711--1721, Lisbon,
  Portugal, 2015.

\bibitem[Williams(1959)]{williams_regression_1959}
Williams, Evan~James.
\newblock \emph{Regression analysis}.
\newblock Wiley, New York, NY, USA, 1959.

\bibitem[Xie et~al.(2016)Xie, Avati, Arivazhagan, Jurafsky, and
  Ng]{xie_neural_2016}
Xie, Ziang, Avati, Anand, Arivazhagan, Naveen, Jurafsky, Dan, and Ng, Andrew~Y.
\newblock Neural {Language} {Correction} with {Character}-{Based} {Attention}.
\newblock \emph{arXiv:1603.09727 [cs]}, March 2016.

\bibitem[Young et~al.(2010)Young, Ga\v{s}i\'c, Keizer, Mairesse, Schatzmann,
  Thomson, and Yu]{young_hidden_2010}
Young, Steve, Ga\v{s}i\'c, Milica, Keizer, Simon, Mairesse, François,
  Schatzmann, Jost, Thomson, Blaise, and Yu, Kai.
\newblock The {Hidden} {Information} {State} model: {A} practical framework for
  {POMDP}-based spoken dialogue management.
\newblock \emph{Computer Speech \& Language}, 24\penalty0 (2):\penalty0
  150--174, April 2010.

\end{thebibliography}
\bibliographystyle{icml2017}

\end{document}